# *Interval-valued q-Rung Orthopair Fuzzy Choquet Integral Operators and Its Application in Group Decision Making*

Benting Wan [1,*], Juelin Huang [2] and Xi Chen [3]

[1] School of software and Internet of things Engineering, Jiangxi University of Finance and Economics, Nanchang 330013, China; wanbenting@jxufe.edu.cn
[2] School of software and Internet of things Engineering, Jiangxi University of Finance and Economics, Nanchang 330013, China; 2202021682@stu.jxufe.edu.cn
[3] School of software and Internet of things Engineering, Jiangxi University of Finance and Economics, Nanchang 330013, China; 517060730@qq.com.
* Correspondence: e-mail: wanbenting@jxufe.edu.cn

**Abstract:** It is more flexible for decision makers to evaluate by interval-valued q-rung orthopair fuzzy set (IVq-ROFS), which offers fuzzy decision-making more applicational space. Meanwhile, Choquet integral uses non-additive set function (fuzzy measure) to describe the interaction between attributes directly. In particular, there are a large number of practical issues that have relevance between attributes. Therefore, this paper proposes the correlation operator and group decision-making method based on the interval-valued q-rung orthopair fuzzy set Choquet integral. First, interval-valued q-rung orthopair fuzzy Choquet integral average operator (IVq-ROFCA) and interval-valued q-rung orthopair fuzzy Choquet integral geometric operator (IVq-ROFCG) are investigated, and their basic properties are proved. Furthermore, several operators based on IVq-ROFCA and IVq-ROFCG are developed. Then, a group decision-making method based on IVq-ROFCA is developed, which can solve the decision-making problems with interaction between attributes. Finally, through the implementation of the warning management system for hypertension, it is shown that the operator and group decision-making method proposed in this paper can handle complex decision-making cases in reality, and the decision result is consistent with the doctor's diagnosis result. Moreover, the comparison with the results of other operators shows that the proposed operators and group decision-making method are correct and effective, and the decision result will not be affected by the change of q value.

## 1. Introduction

The fuzzy set (FS) theory and method proposed by Zadeh [1] have been widely used in real life, such as medical treatment, manufacturing, education, etc. With the increase of people's awareness of complex issues and uncertainty issues, the research on fuzzy theories and methods has received great attention from researchers [2-6]. Since decision-makers in real life need to deal with the possibilities of support, opposition, and neutrality, Atanassov further proposed the intuitionistic fuzzy set (IFS) [7], that is, the sum of membership degree (u) and non-membership degree (v) of each ordered pair is less than or equal to one: $u + v \leq 1$. In response to the sum of membership degree and non-membership degree of each ordered pair being greater than one, Yager proposed the Pythagorean fuzzy set (PFS) [8-9], the sum of the square of membership and membership is less than or equal to one: $u^2 + v^2 \leq 1$, which covers a wider range compared with intuitionistic fuzzy. To deal with much more complicated problems, Yager proposed the situation that the degree of membership and non-membership of the q power is less than or equal to one in 2016: $u^q + v^q \leq 1(q \geq 1)$ [10]. It not only combines IFS and PFS, but also expands the fuzzy set to a wider range of applications based on the different values of q. Researchers have put forward numerous excellent results in recent years: Yager [11-12] and Alajlan studied some properties of q-rung orthopair fuzzy set (q-ROFS) [13]; Liu and Wang proposed q-rung orthopair fuzzy numbers (q-ROFNs) and their properties, and developed q-rung orthopair fuzzy weighted average operators and Bonferroni mean operator, which was used to deal with multi-attribute group decision-making(MAGDM) problems [14-15]; Xing et al. developed a q-rung orthopair fuzzy point weighted aggregation operator and applied it to multi-attribute decision-making [16]; Garg solved the MAGDM issues through the investigated trigonometric operation and connection number based on q-ROFS[17-18]; Hussain et al. also published many research results based on q-ROFNs [19].

However, the attributes are not independent of each other in decision-making, there are more mutual influences and correlations that can be appropriately solved by Choquet integral [20]. The Choquet integral based on fuzzy measures comes decision-making problems related to attributes in handy [21-23], which is had conducted in-depth research: Tan defined the intuitionistic fuzzy Choquet integral average operator and the interval-valued integral average operator [24-26]; Xu investigated the intuitionistic fuzzy Choquet integral average operator and geometric operator [27]. Inspired by Tan and Xu, many scholars expanded the intuitionistic fuzzy Choquet integral operator. Besides, Xu proposed intuitionistic fuzzy weighted average and weighted geometric operators [28-29], Zhao et al. proposed the q-rung intuitionistic fuzzy operators [30], and Tan presented the q-rung intuitionistic fuzzy geometric operators [31], which provide the theoretical basis for the subsequent study of q-ROFSs. Many experts and scholars have conducted a lot of studies on fuzzy integral (including Choquet integral) group decision-making methods [32-33], and found that the group decision-making methods can effectively process decision data of multiple attributes and multiple experts. Xing et al. [34] studied Choquet integral based on q-rung orthopair and proposed related MAGDM methods; Keikha et al. [35] combined the Choquet integral and TOPSIS method to solve the problem; Teng et al. [36] developed the generalized Shapley probabilistic linguistic Choquet average (GS-PLCA) operator and investigated a method that can deal with large group decision-making (LGDM) issues. Then, extending the Choquet integral and combining it with the group decision-making method for in-depth research is of practical significance for solving many complex and uncertain problems in real decision-making.

Inspired by Choquet integral operator and q-ROFS, this paper analyses the MAGDM problem based on IVq-ROFS, proposes IVq-ROFCA, IVq-ROFCG, then proves their properties and develops some weighted operators based them. Combined with the group decision-making method proposed in this paper, the case of warning management system for hypertension is verified and compared. The research results of this paper are as follows.

(1) Based on the Choquet integral and q-ROFNs, this paper proposes IVq-ROFCA, IVq-ROFCG, including some properties and extended operators of them. In addition, these operators are verified, and idempotency, boundedness, commutativity and monotonicity of them are proved;

(2) A group decision-making method is proposed on the basis of IVq-ROFCA. In this method, two aggregations of experts and attributes are carried out by IVq-ROFCA respectively, which collects IVq-ROFNs s corresponding to each alternative. The ranking result of alternatives is obtained via the score function and exact function of IVq-ROFNs.

(3) We also apply the proposed method into the case of warning management system for hypertension, and the decision result correctly satisfies the diagnosis of doctor. It is not only found that the decision results are discussed with methods proposed by other papers, but when q takes the values 2, 3, 4, 5, same decision results can be also derived.

The organization of this paper is as follows: Section 2 reviews the concept of IVq-ROFS and Choquet integral operator; Section 3 proposes the IVq-ROFCA, IVq-ROFCG; Moreover, Section 4 presents the group decision-making method based on given operators; Section 5 applies solves practical cases by proposed group decision-making method and provides comparative analysis. Section 6 concludes this paper.

## 2. Preliminaries

In this section, we make a brief review of IVq-ROFS and Choquet integral operator.

**Definition 1 [11].** Let $X = \{x_1, x_2, \ldots, x_n\}$ be a fixed set, $\tilde{a} = \{\langle x_i, t_{\tilde{a}}(x_i), f_{\tilde{a}}(x_i)\rangle | x_i \in X\}$ is a q-ROFS, where $t_{\tilde{a}}: X \in [0,1]$, $f_{\tilde{a}}: X \in [0,1]$, satisfy Equation (1):

$$0 \leq (t_{\tilde{a}}(x_i))^q + (f_{\tilde{a}}(x_i))^q \leq 1 \tag{1}$$

Where, $q \geq 1$, and for all $x_i \in X$, $t_{\tilde{a}}(x_i)$ are the degree of membership; $f_{\tilde{a}}(x_i)$ is the degree of non-membership, and the degree of hesitation $\pi_{\tilde{a}}(x_i)$ can be expressed as Equation (2):

$$\pi_{\tilde{a}}(x_i) = \sqrt[q]{1 - t_{\tilde{a}}(x_i)^q - f_{\tilde{a}}(x_i)^q} (q \geq 1) \tag{2}$$

**Definition 2 [37].** Given a fixed Set $X = \{x_1, x_2, \ldots, x_n\}$, IVq-ROFS $a$ on $X$ is defined as shown in Equation (3):

$$a = \{< x_i, t_a(x_i), f_a(x_i) > | x_i \in X\} \tag{3}$$

The membership is represented by interval values, satisfies: $t_a(x_i) = [t_a^-(x_i), t_a^+(x_i)] \subseteq [0,1]$, the non-membership satisfies: $f_a(x_i) = [f_a^-(x_i), f_a^+(x_i)] \subseteq [0,1]$, and also satisfied: $0 \leq (t_a^+(x_i))^q + (f_a^+(x_i))^q \leq 1, (q \geq 1)$. The degree of hesitation of $a$ is shown as Equation (4).

$$\pi_a(x_i) = [\pi_a^-(x_i), \pi_a^+(x_i)] =$$
$$[\sqrt[q]{1-(t_a^+(x_i))^q-(f_a^+(x_i))^q}, \sqrt[q]{1-(t_a^-(x_i))^q-(f_a^-(x_i))^q}] \tag{4}$$

In particular, when $q = 1$, IVq-ROFS reduces to interval-valued intuitionistic fuzzy set (IVIFS).

**Definition 3 [38].** Let $a_1 = ([t_{a_1}^-, t_{a_1}^+], [f_{a_1}^-, f_{a_1}^+])$ and $a_2 = ([t_{a_2}^-, t_{a_2}^+], [f_{a_2}^-, f_{a_2}^+])$ are interval valued q-rung orthopair numbers (IVq-ROFNs), $q \geq 1$, Then the Equation (5), (6), (7) and (8) are established.

$$a_1 \oplus a_2 = <\left[\sqrt[q]{(t_{a_1}^-)^q + (t_{a_2}^-)^q - (t_{a_1}^-)^q(t_{a_2}^-)^q}, \sqrt[q]{(t_{a_1}^+)^q + (t_{a_2}^+)^q - (t_{a_1}^+)^q(t_{a_2}^+)^q}\right],$$
$$[f_{a_1}^- f_{a_2}^-, f_{a_1}^+ f_{a_2}^+] > \tag{5}$$

$$a_1 \otimes a_2 = <[t_{a_1}^- t_{a_2}^-, t_{a_1}^+ t_{a_2}^+],$$
$$[\sqrt[q]{(f_{a_1}^-)^2 + (f_{a_2}^-)^2 - (f_{a_1}^-)^2(f_{a_2}^-)^2}, \sqrt[q]{(f_{a_1}^+)^2 + (f_{a_2}^+)^2 - (f_{a_1}^+)^2(f_{a_2}^+)^2}] > \tag{6}$$

$$\lambda a_1 = <\left[\sqrt[q]{1-(1-(t_{a_1}^-)^q)^\lambda}, \sqrt[q]{1-(1-(t_{a_1}^+)^q)^\lambda}\right], [(f_{a_1}^-)^\lambda, (f_{a_1}^+)^\lambda] > \tag{7}$$

$$a_1^\lambda = <[(t_{a_1}^-)^\lambda, (t_{a_1}^+)^\lambda], \left[\sqrt[q]{1-(1-(f_{a_1}^-)^q)^\lambda}, \sqrt[q]{1-(1-(f_{a_1}^+)^q)^\lambda}\right] > \tag{8}$$

**Definition 4 [38].** For any IVq-ROFNs $a = ([t_a^-, t_a^+], [f_a^-, f_a^+])$, the score function is shown as Equation (9).

$$S(a) = \frac{1}{2}[(t_a^-)^q + (t_a^+)^q - (f_a^-)^q - (f_a^+)^q], (q \geq 1) \tag{9}$$

**Definition 5 [38].** For any IVq-ROFNs $a = ([t_a^-, t_a^+], [f_a^-, f_a^+])$, the exact function is shown as Equation (10).

$$H(a) = \frac{1}{2}[(t_a^-)^q + (t_a^+)^q + (f_a^-)^q + (f_a^+)^q](q \geq 1) \tag{10}$$

**Definition 6 [38].** For any two IVq-ROFNs $a_1$ and $a_2$, the comparison law is as follows:
  If $S(a_1) > S(a_2)$, then $a_1 > a_2$;
  If $S(a_1) < S(a_2)$, then $a_1 < a_2$;
  If $S(a_1) = S(a_2)$:
    if $H(a_1) > H(a2)$, then $a_1 > a_2$;
    if $H(a_1) < H(a_2)$, then $a_1 < a_2$;
    if $H(a_1) = H(a_2)$, then $a_1 = a_2$.

**Definition 7 [20].** Let $X = \{x_1, x_2, \ldots, x_n\}$ be a universe of discourse, $f$ be a positive real-valued function and $\mu$ be the fuzzy measure on $X$. Then the discrete Choquet integral of $f$ on fuzzy measure $\mu$ is defined as Equation (11).

$$\int f d\mu = \sum_{i=1}^n f(x_{\sigma(i)})[\mu(B_{\sigma(i)}) - \mu(B_{\sigma(i-1)})] \tag{11}$$

Where $(\sigma(1), \sigma(2), \cdots, \sigma(n))$ is a permutation of $(1, 2, \ldots n)$ that satisfies $f(x_{\sigma(1)}) \geq f(x_{\sigma(2)}) \geq \cdots \geq f(x_{\sigma(n)})$, $B_{\sigma(I)} = \{x_{\sigma(1)}, x_{\sigma(2)}, \ldots, x_{\sigma(i)}\}$, $i = 1, 2, \ldots, n, B_{\sigma(0)} = \emptyset$.

## 3. Interval-valued q-rung orthopair fuzzy Choquet integral operators

In this section, we investigate IVq-ROFCA and IVq-ROFCG operators, then extend their weighted operators and some properties.

### 3.1. IVq-ROFCA

**Definition 8.** Let $\mu$ be the fuzzy measure on nonempty finite set $X = \{x_1, x_2, \ldots, x_n\}$ ($\mu(\emptyset) = 0$), and $a(x_i) = <[t_a^-(x_i), t_a^+(x_i)], [f_a^-(x_i), f_a^+(x_i)]> (i = 1, 2, \cdots, n)$ be IVq-ROFNs; then the IVq-ROFCA can be defined as Equation (12).

$$(C_1)\int a d\mu = IVq - ROFCA(a(x_1), a(x_2), \ldots, a(x_n)) =$$
$$\sum_{i=1}^n [\mu(B_{\sigma(i)}) - \mu(B_{\sigma(i-1)})] a(x_{\sigma(i)}) \tag{12}$$

Where $(C_1)\int \alpha d\mu$ represemates the Choquet integral, $(\sigma(1), \sigma(2),\cdots,\sigma(n))$ is a permutation of $(1,2,...n)$ that satisfies $a(x_{\sigma(1)}) \geq a(x_{\sigma(2)}) \geq \cdots \geq a(x_{\sigma(n)})$, $B_{\sigma(I)} = \{x_{\sigma(1)}, x_{\sigma(2)},...,x_{\sigma(i)}\}$, $i = 1,2,...,n$, $B_{\sigma(0)} = \emptyset$.

**Theorem 1.** If there are fuzzy measure $\mu$ ( $\mu(\emptyset) = 0$ ) and IVq-ROFNs $a(x_i) = <[t_a^-(x_i), t_a^+(x_i)], [f_a^-(x_i), f_a^+(x_i)]>$ ($i = 1, 2, ..., n$). Then the IVq-ROFCA can be derived as Equation (13):

$$IVq - ROFCA(a(x_1), a(x_2), ..., a(x_n)) =$$

$$< \left[ \sqrt[q]{1 - \prod_{i=1}^{n}(1 - t_a^-(x_{\sigma(i)})^q)^{\mu(B_{\sigma(i)}) - \mu(B_{\sigma(i-1)})}}, \sqrt[q]{1 - \prod_{i=1}^{n}(1 - t_a^+(x_{\sigma(i)})^q)^{\mu(B_{\sigma(i)}) - \mu(B_{\sigma(i-1)})}} \right], \left[ \prod_{i=1}^{n}(f_a^-(x_{\sigma(i)}))^{\mu(B_{\sigma(i)}) - \mu(B_{\sigma(i-1)})}, \prod_{i=1}^{n}(f_a^+(x_{\sigma(i)}))^{\mu(B_{\sigma(i)}) - \mu(B_{\sigma(i-1)})} \right] > \quad (13)$$

**Proof of Theorem 1:** We prove Equation (13) by mathematical induction.
If n=2,

IVq-ROFCA$(a(x_1), a(x_2))$

$$= \prod_{i=1}^{2}[\mu(B_{\sigma(i)}) - \mu(B_{\sigma(i-1)})]a(x_{\sigma(i)})$$

$$= < \left[ \sqrt[q]{1 - \prod_{i=1}^{2}(1 - t_a^-(x_{\sigma(i)})^q)^{\mu(B_{\sigma(i)}) - \mu(B_{\sigma(i-1)})}}, \sqrt[q]{1 - \prod_{i=1}^{2}(1 - t_a^+(x_{\sigma(i)})^q)^{\mu(B_{\sigma(i)}) - \mu(B_{\sigma(i-1)})}} \right], \left[ \prod_{i=1}^{2}(f_a^-(x_{\sigma(i)}))^{\mu(B_{\sigma(i)}) - \mu(B_{\sigma(i-1)})}, \prod_{i=1}^{2}(f_a^+(x_{\sigma(i)}))^{\mu(B_{\sigma(i)}) - \mu(B_{\sigma(i-1)})} \right] >$$

If n=k,

IVq-ROFCA$(a(x_1), a(x_2), \cdots, a(x_k))$

$$= < \left[ \sqrt[q]{1 - \prod_{i=1}^{k}(1 - t_a^-(x_{\sigma(i)})^q)^{\mu(B_{\sigma(i)}) - \mu(B_{\sigma(i-1)})}}, \sqrt[q]{1 - \prod_{i=1}^{k}(1 - t_a^+(x_{\sigma(i)})^q)^{\mu(B_{\sigma(i)}) - \mu(B_{\sigma(i-1)})}} \right], \left[ \prod_{i=1}^{k}(f_a^-(x_{\sigma(i)}))^{\mu(B_{\sigma(i)}) - \mu(B_{\sigma(i-1)})}, \prod_{i=1}^{k}(f_a^+(x_{\sigma(i)}))^{\mu(B_{\sigma(i)}) - \mu(B_{\sigma(i-1)})} \right] >$$

If n=k+1, the results of IVq-ROFCA are as follows:

IVq-ROFCA$(a(x_1), a(x_2), \cdots, a(x_{k+1}))$

$$= \prod_{i=1}^{k+1}[\mu(B_{\sigma(i)}) - \mu(B_{\sigma(i-1)})]a(x_{\sigma(i)})$$

$$= \prod_{i=1}^{k}[\mu(B_{\sigma(i)}) - \mu(B_{\sigma(i-1)})]a(x_{\sigma(i)}) \oplus [\mu(B_{\sigma(k+1)}) - \mu(B_{\sigma(k)})]a(x_{\sigma(k+1)})$$

$$= < \left[ \sqrt[q]{1 - \prod_{i=1}^{k}(1 - t_a^-(x_{\sigma(i)})^q)^{\mu(B_{\sigma(i)}) - \mu(B_{\sigma(i-1)})}}, \sqrt[q]{1 - \prod_{i=1}^{k}(1 - t_a^+(x_{\sigma(i)})^q)^{\mu(B_{\sigma(i)}) - \mu(B_{\sigma(i-1)})}} \right], \left[ \prod_{i=1}^{k}(f_a^-(x_{\sigma(i)}))^{\mu(B_{\sigma(i)}) - \mu(B_{\sigma(i-1)})}, \prod_{i=1}^{k}(f_a^+(x_{\sigma(i)}))^{\mu(B_{\sigma(i)}) - \mu(B_{\sigma(i-1)})} \right] > \oplus$$

$$< \left[ \begin{array}{l} \sqrt[q]{1 - (1 - t_a^-(x_{\sigma(k+1)})^q)^{\mu(B_{\sigma(k+1)}) - \mu(B_{\sigma(k)})}}, \\ \sqrt[q]{1 - (1 - t_a^+(x_{\sigma(k+1)})^q)^{\mu(B_{\sigma(k+1)}) - \mu(B_{\sigma(k)})}} \end{array} \right], \left[ \begin{array}{l} (f_a^-(x_{\sigma(k+1)}))^{\mu(B_{\sigma(k+1)}) - \mu(B_{\sigma(k)})}, \\ (f_a^+(x_{\sigma(k+1)}))^{\mu(B_{\sigma(k+1)}) - \mu(B_{\sigma(k)})} \end{array} \right] >$$

$$= < \left[ \begin{array}{l} \sqrt[q]{1 - \prod_{i=1}^{k+1}(1 - t_a^-(x_{\sigma(i)})^q)^{\mu(B_{\sigma(i)}) - \mu(B_{\sigma(i-1)})}}, \\ \sqrt[q]{1 - \prod_{i=1}^{k+1}(1 - t_a^+(x_{\sigma(i)})^q)^{\mu(B_{\sigma(i)}) - \mu(B_{\sigma(i-1)})}} \end{array} \right], \left[ \begin{array}{l} \prod_{i=1}^{k+1}(f_a^-(x_{\sigma(i)}))^{\mu(B_{\sigma(i)}) - \mu(B_{\sigma(i-1)})}, \\ \prod_{i=1}^{k+1}(f_a^+(x_{\sigma(i)}))^{\mu(B_{\sigma(i)}) - \mu(B_{\sigma(i-1)})} \end{array} \right] > \square$$

The Equation (13) holds, the Theorem 1 is established.

**Example 1.** There are three suppliers $\{x_1, x_2, x_3\}$ to be chosen. The core competitiveness of suppliers can be evaluated by three criteria $\{C_1, C_2, C_3\}$: $C_1$: the level of technological innovation; $C_2$: the ability of circulation control; $C_3$: the capability of management. The decision matrix $A = (a_{ij})_{3 \times 3}$ is generated by experts' evaluation, in which the evaluation value $a_{ij} = < [t_{a_{ij}}^-, t_{a_{ij}}^+], [f_{a_{ij}}^-, f_{a_{ij}}^+] > (i, j = 1,2,3)$ is IVq-ROFNs, shown in Table 1. Now it needs to evaluate the core competitiveness of $\{x_1, x_2, x_3\}$ according to the decision matrix $A = (a_{ij})_{3 \times 3}$, which provides the reference for the manufacturing enterprise to select. The fuzzy measure $\mu$ represents each attribute and attributes set as follow:

$$\mu(\emptyset) = 0, \mu(\{C_1, C_2, C_3\}) = 1, \mu(\{C_1\}) = \mu(\{C_2\}) = 0.4, \mu(\{C_3\}) = 0.3,$$

$$\mu(\{C_1, C_2\}) = 0.5, \mu(\{C_1, C_3\}) = \mu(\{C_2, C_3\}) = 0.8$$

**Table 1.** decision matrix $A = (a_{ij})_{3 \times 3}$

|       | $C_1$                    | $C_2$                    | $C_3$                    |
|-------|--------------------------|--------------------------|--------------------------|
| $x_1$ | <[0.7,0.9],[0.3,0.5]>    | <[0.3,0.4],[0.5,0.6]>    | <[0.5,0.6],[0.4,0.5]>    |
| $x_2$ | <[0.6,0.8],[0.4,0.5]>    | <[0.3,0.5],[0.5,0.7]>    | <[0.4,0.7],[0.3,0.5]>    |
| $x_3$ | <[0.6,0.8],[0.3,0.4]>    | <[0.4,0.6],[0.4,0.5]>    | <[0.5,0.7],[0.3,0.5]>    |

When q is 2, the evaluation $C_1$ of supplier $x_1$ dissatisfies $t_{a_{11}}^{+\,q} + f_{a_{11}}^{+\,q} \leq 1$; When q is 3, all elements in $A$ meet $t_{a_{ij}}^+ + f_{a_{ij}}^+ \leq 1$. Therefore, set the value of q as 3, and the results of IVq-ROFCA are as follows:

$$r_1 = IVq - ROFCA(x_1(C_1), x_1(C_2), x_1(C_3)) = <[0.57, 0.78], [0.37, 0.52]>$$
$$r_2 = IVq - ROFCA(x_2(C_1), x_2(C_2), x_2(C_3)) = <[0.49, 0.73], [0.37, 0.46]>$$
$$r_3 = IVq - ROFCA(x_3(C_1), x_3(C_2), x_3(C_3)) = <[0.50, 0.70], [0.32, 0.46]>$$

It's easy to find $r_1$, $r_2$ and $r_3$ are IVq-ROFNs. According to score function that shows in Equation (9), the order of their corresponding score function can be derived: $S(r_1) > S(r_2) > S(r_3)$, which represents that supplier $x_1$ is better than $x_2$ and $x_3$.

**Theorem 2.** Suppose that $a(x_i) = < [t_a^-(x_i), t_a^+(x_i)], [f_a^-(x_i), f_a^+(x_i)] > (i = 1,2,...,n)$ is a set of IVq-ROFNs, $\mu$ is the fuzzy measure on a nonempty finite set $X$, then the IVq-ROFCA have the following four properties:

(1) Idempotency: If it is existed $a(x) = < [t_a^-(x), t_a^+(x)], [f_a^-(x), f_a^+(x)] > (i = 1,2,...,n)$, and $a(x_i) = a(x)$ for all elements in $a(x_i)$. Then the following equation can be obtained: IVq-ROFCA$(a(x_1), a(x_2), ..., a(x_n)) = a(x)$.

(2) Boundedness: Assume $a_{min} = <[min(t_{a_i}^-), min(t_{a_i}^+)], [max(f_{a_i}^-), max(f_{a_i}^+)] >$ and $a_{max} = <[max(t_{a_i}^-), max(t_{a_i}^+)], [min(f_{a_i}^-), min(f_{a_i}^+)] >$ are two IVq-ROFNs in $a(x_i)$, then $a_{min} \leq$ IVq-ROFCA$(a(x_1), a(x_2), ..., a(x_n)) \leq a_{max}$.

(3) Commutativity: Suppose that $(a'(x_1), a'(x_2), ..., a'(x_n))$ is any permutation of $(a(x_1), a(x_2), ..., a(x_n))$, then the equality IVq-ROFCA$(a(x_1), a(x_2), ..., a(x_n)) =$ IVq-ROFCA$(a'(x_1), a'(x_2), ..., a'(x_n))$ holds.

(4) Monotonicity: Let $\beta(x_i) = < [t_\beta^-(x_i), t_\beta^+(x_i)], [f_\beta^-(x_i), f_\beta^+(x_i)] > (i = 1,2,...,n)$ be a IVq-ROFS. For any $i$, which satisfies that $t_{a_i}^- \leq t_{\beta_i}^-, t_{a_i}^+ \leq t_{\beta_i}^+, f_{a_i}^- \geq f_{\beta_i}^-, f_{a_i}^+ \geq f_{\beta_i}^+$, then IVq-ROFCA$(a(x_1), a(x_2), \cdots, a(x_n)) \leq$ IVq-ROFCA$(\beta(x_1), \beta(x_2), \cdots, \beta(x_n))$.

The proof process of Theorem is given below.

**Proof of Theorem 2:**

(1) Because all elements in the $a(x_i)$ are equal and $a(x_i) = a(x)$,

$$\mu(x_i) = \frac{1}{n}, \text{ and } \mu(B_{\sigma(i)}) - \mu(B_{\sigma(i-1)}) = \frac{1}{n}$$

$$IVq - ROFCA(a(x_1), a(x_2), ..., a(x_n)) =$$

$$< \left[ \begin{array}{c} \sqrt[q]{1 - \prod_{i=1}^{n}(1 - t_a^-(x_{\sigma(i)})^q)^{\mu(B_{\sigma(i)})-\mu(B_{\sigma(i-1)})}}, \\ \sqrt[q]{1 - \prod_{i=1}^{n}(1 - t_a^+(x_{\sigma(i)})^q)^{\mu(B_{\sigma(i)})-\mu(B_{\sigma(i-1)})}} \end{array} \right], \left[ \begin{array}{c} \prod_{i=1}^{n}(f_a^-(x_{\sigma(i)}))^{\mu(B_{\sigma(i)})-\mu(B_{\sigma(i-1)})}, \\ \prod_{i=1}^{n}(f_a^+(x_{\sigma(i)}))^{\mu(B_{\sigma(i)})-\mu(B_{\sigma(i-1)})} \end{array} \right] >$$

$$=< \left[ \begin{array}{c} \sqrt[q]{1 - \prod_{i=1}^{n}(1 - t_a^-(x)^q)^{\frac{1}{n}}}, \\ \sqrt[q]{1 - \prod_{i=1}^{n}(1 - t_a^+(x)^q)^{\frac{1}{n}}} \end{array} \right], \left[ \begin{array}{c} \prod_{i=1}^{n}(f_a^-(x))^{\frac{1}{n}}, \\ \prod_{i=1}^{n}(f_a^+(x))^{\frac{1}{n}} \end{array} \right] >$$

$$=< \left[ \begin{array}{c} \sqrt[q]{1 - (1 - t_a^-(x)^q)}, \\ \sqrt[q]{1 - (1 - t_a^+(x)^q)} \end{array} \right], \left[ \begin{array}{c} f_a^-(x), \\ f_a^+(x) \end{array} \right] >$$

$$=< \left[ \begin{array}{c} \sqrt[q]{t_a^-(x)^q}, \\ \sqrt[q]{t_a^+(x)^q} \end{array} \right], \left[ \begin{array}{c} f_a^-(x), \\ f_a^+(x) \end{array} \right] >= a(x) \square$$

(2) According to the Equation (9), it can be seen that:

$$S(a_{min}) = \frac{1}{2}[(min\ (t_{a_i}^-))^q + (min\ (t_{a_i}^+))^q - (max\ (f_{a_i}^-))^q - (max(f_{a_i}^+))^q]$$

$$S(a_{max}) = \frac{1}{2}[(max\ (t_{a_i}^-))^q + (max\ (t_{a_i}^+))^q - (min\ (f_{a_i}^-))^q - (min(f_{a_i}^+))^q]$$

For any $t_{a_i}^-$, it satisfies:

$$(min\ (t_{a_i}^-))^q \leq (t_{a_i}^-)^q \leq (max\ (t_{a_i}^-))^q$$
$$1 - (min\ (t_{a_i}^-))^q \geq 1 - (t_{a_i}^-)^q \geq 1 - (max\ (t_{a_i}^-))^q$$

$$\prod_{i=1}^{n}(1 - (min\ (t_{a_i}^-))^q)^{\mu(B_{\sigma(i)})-\mu(B_{\sigma(i-1)})} \geq \prod_{i=1}^{n}(1 - (t_{a_i}^-)^q)^{\mu(B_{\sigma(i)})-\mu(B_{\sigma(i-1)})}$$
$$\geq \prod_{i=1}^{n}(1 - (max\ (t_{a_i}^-))^q)^{\mu(B_{\sigma(i)})-\mu(B_{\sigma(i-1)})}$$

$$1 - \prod_{i=1}^{n}(1 - (min\ (t_{a_i}^-))^q)^{\mu(B_{\sigma(i)})-\mu(B_{\sigma(i-1)})} \leq 1 - \prod_{i=1}^{n}(1 - (t_{a_i}^-)^q)^{\mu(B_{\sigma(i)})-\mu(B_{\sigma(i-1)})}$$
$$\leq 1 - \prod_{i=1}^{n}(1 - (max\ (t_{a_i}^-))^q)^{\mu(B_{\sigma(i)})-\mu(B_{\sigma(i-1)})}$$

Namely:

$$(min\ (t_{a_i}^-))^q \leq 1 - \prod_{i=1}^{n}(1 - (t_{a_i}^-)^q)^{\mu(B_{\sigma(i)})-\mu(B_{\sigma(i-1)})} \leq (max\ (t_{a_i}^-))^q$$

Similarly:

$$(min(t_{a_i}^+))^q \leq -\prod_{i=1}^{n}(1 - (t_{a_i}^+)^q)^{\mu(B_{\sigma(i)})-\mu(B_{\sigma(i-1)})} \leq (max(t_{a_i}^+))^q$$

$$(min(f_{a_i}^-))^q \leq \prod_{i=1}^{n}((f_{a_i}^-)^q)^{\mu(B_{\sigma(i)})-\mu(B_{\sigma(i-1)})} \leq (max(f_{a_i}^-))^q$$

$$(min(f_{a_i}^+))^q \leq \prod_{i=1}^{n}((f_{a_i}^-)^q)^{\mu(B_{\sigma(i)})-\mu(B_{\sigma(i-1)})} \leq (max(f_{a_i}^+))^q$$

Sorted out:

$$S(a_{min}) \leq S(IVq - ROFCA(a_1, a_2, \ldots, a_n)) \leq S(a_{max})$$
$$a_{min} \leq IVq - ROFCA(a_1, a_2, \ldots, a_n) \leq a_{max} \square$$

(3) According to Definition 8, it can be easily derived that

$$IVq - ROFCA(a(x_1), a(x_2), \ldots, a(x_n)) =$$

$$< \left[ \sqrt[q]{1 - \prod_{i=1}^{n}(1 - t_a^-(x_{\sigma(i)})^q)^{\mu(B_{\sigma(i)}) - \mu(B_{\sigma(i-1)})}}, \sqrt[q]{1 - \prod_{i=1}^{n}(1 - t_a^+(x_{\sigma(i)})^q)^{\mu(B_{\sigma(i)}) - \mu(B_{\sigma(i-1)})}} \right], \left[ \prod_{i=1}^{n}(f_a^-(x_{\sigma(i)}))^{\mu(B_{\sigma(i)}) - \mu(B_{\sigma(i-1)})}, \prod_{i=1}^{n}(f_a^+(x_{\sigma(i)}))^{\mu(B_{\sigma(i)}) - \mu(B_{\sigma(i-1)})} \right] >$$

$= \text{IVq-ROFCA}(a'(x_1), a'(x_2), \ldots, a'(x_n))\square$

(4) From the Definition 8 and $t_{a_i}^- \leq t_{\beta_i}^-$, it can be found that

$$t_{a_i}^-(x_{\sigma(i)})^q \leq t_{\beta_i}^-(x_{\sigma(i)})^q$$
$$1 - t_{a_i}^-(x_{\sigma(i)})^q \geq 1 - t_{\beta_i}^-(x_{\sigma(i)})^q$$

Thus,

$$\prod_{i=1}^{n}(1 - t_{a_i}^-(x_{\sigma(i)})^q)^{\mu(B_{\sigma(i)}) - \mu(B_{\sigma(i-1)})} \geq \prod_{i=1}^{n}(1 - t_{\beta_i}^-(x_{\sigma(i)})^q)^{\mu(B_{\sigma(i)}) - \mu(B_{\sigma(i-1)})}$$

$$\sqrt[q]{1 - \prod_{i=1}^{n}(1 - t_{a_i}^-(x_{\sigma(i)})^q)^{\mu(B_{\sigma(i)}) - \mu(B_{\sigma(i-1)})}} \leq \sqrt[q]{1 - \prod_{i=1}^{n}(1 - t_{\beta_i}^-(x_{\sigma(i)})^q)^{\mu(B_{\sigma(i)}) - \mu(B_{\sigma(i-1)})}}$$

Similarly,

$$\sqrt[q]{1 - \prod_{i=1}^{n}(1 - t_{a_i}^+(x_{\sigma(i)})^q)^{\mu(B_{\sigma(i)}) - \mu(B_{\sigma(i-1)})}} \leq \sqrt[q]{1 - \prod_{i=1}^{n}(1 - t_{\beta_i}^+(x_{\sigma(i)})^q)^{\mu(B_{\sigma(i)}) - \mu(B_{\sigma(i-1)})}}$$

$$\prod_{i=1}^{n}(f_{a_i}^-(x_{\sigma(i)}))^{\mu(B_{\sigma(i)}) - \mu(B_{\sigma(i-1)})} \geq \prod_{i=1}^{n}(f_{\beta_i}^-(x_{\sigma(i)}))^{\mu(B_{\sigma(i)}) - \mu(B_{\sigma(i-1)})}$$

$$\prod_{i=1}^{n}(f_{a_i}^+(x_{\sigma(i)}))^{\mu(B_{\sigma(i)}) - \mu(B_{\sigma(i-1)})} \geq \prod_{i=1}^{n}(f_{\beta_i}^+(x_{\sigma(i)}))^{\mu(B_{\sigma(i)}) - \mu(B_{\sigma(i-1)})}$$

According to Definition 6, it is obviously discovered that $\text{IVq-ROFCA}(a(x_1), a(x_2), \cdots, a(x_n)) \leq \text{IVq-ROFCA}(\beta(x_1), \beta(x_2), \cdots, \beta(x_n))$ $\square$.

**Definition 9.** Let $\mu$ be a fuzzy measure on $X$ ($\mu(\emptyset) = 0$) and $a(x_i) = <[t_a^-(x_i), t_a^+(x_i)], [f_a^-(x_i), f_a^+(x_i)]> (i = 1,2,\cdots,n)$ be n IVq-ROFNs.

(1) If $\mu(B \cup C) = \mu(B) + \mu(C)$ for all $B, C \subseteq X$, $B \cap C = \emptyset$, and it is independent for any elements in $X$ that means $\mu(B) = \sum_{x_i \in B} \mu(\{x_i\})$. Then, the IVq-ROFCA transforms to interval-valued q-rung orthopair fuzzy weighted Choquet averaging (IVq-ROFWCA) operator, which shows as:

$$IVq - ROFWCA(a(x_1), a(x_2), \ldots, a(x_n)) =$$

$$< \left[ \sqrt[q]{1 - \prod_{i=1}^{n}(1 - t_a^-(x_{\sigma(i)})^q)^{\mu(\{x_i\})}}, \sqrt[q]{1 - \prod_{i=1}^{n}(1 - t_a^+(x_{\sigma(i)})^q)^{\mu(\{x_i\})}} \right], \left[ \prod_{i=1}^{n}(f_a^-(x_{\sigma(i)}))^{\mu(\{x_i\})}, \prod_{i=1}^{n}(f_a^+(x_{\sigma(i)}))^{\mu(\{x_i\})} \right] > \quad (14)$$

(2) If for all $B \subseteq X$, there is $\mu(B) = \sum_{i=1}^{|B|} \lambda_i$, we have $\lambda_i = \mu(B_{\sigma(i)}) - \mu(B_{\sigma(i-1)})(i = 1,2,\ldots,n)$, where $\lambda = (\lambda_1, \lambda_2, \ldots, \lambda_n)^T (\lambda_n \geq 0)$ and $\sum_{i=1}^{n} \lambda_i = 1$ $(i = 1,2,\ldots,n)$. Further, the IVq-ROFCA operator reduce to interval-valued q-rung orthopair fuzzy order Choquet averaging (IVq-ROFOCA) operator as Equation (15).

$$IVq - ROFOCA(a(x_1), a(x_2), \ldots, a(x_n)) =$$

$$< \left[ \sqrt[q]{1 - \prod_{i=1}^{n}(1 - t_a^-(x_{\sigma(i)})^q)^{\lambda_i}}, \sqrt[q]{1 - \prod_{i=1}^{n}(1 - t_a^+(x_{\sigma(i)})^q)^{\lambda_i}} \right], \left[ \prod_{i=1}^{n}(f_a^-(x_{\sigma(i)}))^{\lambda_i}, \prod_{i=1}^{n}(f_a^+(x_{\sigma(i)}))^{\lambda_i} \right] > \quad (15)$$

(3) If for all $B \subseteq X$, there is $\mu(B) = Q(\sum_{x_i \in B} \mu(\{x_i\}))$ where Q is a basic unit-interval monotonic function, satisfies monotonicity in $[0,1]$ and follows properties: $(i) Q(0) = 0; (ii) Q(1) =$

$1; (iii) Q(x) \geq Q(y)$ for $x > y$. Then we let $w_i = \mu(B_{\sigma(i)}) - \mu(B_{\sigma(i-1)}) = Q(\sum_{j \leq i} \mu(\{x_{\sigma(i)}\})) - Q(\sum_{j < i} \mu(\{x_{\sigma(i)}\})) (i = 1,2, \ldots, n)$, where $w = (w_1, w_2, \ldots, w_n)^T (w_n \geq 0)$ and $\sum_{i=1}^{n} w_i = 1 (i = 1,2, \ldots, n)$. The IVq-ROFCA change to Equation (16) that is called interval-valued q-rung orthopair fuzzy order weighted Choquet averaging (IVq-ROFOWCA) operator.

$$IVq - ROFOWCA(a(x_1), a(x_2), \ldots, a(x_n)) =$$

$$< \begin{bmatrix} \sqrt[q]{1 - \prod_{i=1}^{n}(1 - t_a^-(x_{\sigma(i)})^q)^{w_i}}, \\ \sqrt[q]{1 - \prod_{i=1}^{n}(1 - t_a^+(x_{\sigma(i)})^q)^{w_i}} \end{bmatrix}, \left[\prod_{i=1}^{n}(f_a^-(x_{\sigma(i)}))^{w_i}, \prod_{i=1}^{n}(f_a^+(x_{\sigma(i)}))^{w_i}\right] > \quad (16)$$

Especially, if for all $i = 1,2, \ldots, n$, existing that $\mu(\{x_i\}) = \frac{1}{n}$, then the IVq-ROFOWCA operator reduces to the IVq-ROFOCA operator. In addition, we develop the IVq-ROFCG operator next.

### 3.2. IVq-ROFCG

**Definition 10.** Let $\mu$ be a fuzzy measure on on nonempty finite set $X = \{x_1, x_2, \ldots, x_n\}$ ($\mu(\emptyset) = 0$), and $a(x_i) = <[t_a^-(x_i), t_a^+(x_i)], [f_a^-(x_i), f_a^+(x_i)]> (i = 1,2, \ldots, n)$ be IVq-ROFNs. Then, the IVq-ROFCG can be defined as Equation (17):

$$(C_2) \int a \, d\mu = IVq - ROFCG(a(x_1), a(x_2), \ldots, a(x_n)) =$$

$$\prod_{i=1}^{n} [\mu(B_{\sigma(i)}) - \mu(B_{\sigma(i-1)})] a(x_{\sigma(i)}) \quad (17)$$

Where Choquet integral expresses as $(C_2) \int a \, d\mu$, $(\sigma(1), \sigma(2), \ldots, \sigma(n))$ is a permutation of $(1,2, \ldots n)$ which satisfies that $a(x_{\sigma(1)}) \geq a(x_{\sigma(2)}) \geq \cdots \geq a(x_{\sigma(n)})$, $B_{\sigma(i)} = \{x_{\sigma(1)}, x_{\sigma(2)}, \ldots, x_{\sigma(i)}\}, i = 1,2, \ldots, n, B_{\sigma(0)} = \emptyset$.

**Theorem 3.** If $\mu$ is a fuzzy measure and $a(x_i) = <[t_a^-(x_i), t_a^+(x_i)], [f_a^-(x_i), f_a^+(x_i)]> (i = 1,2, \ldots n)$ is IVq-ROFNs, $\mu(\emptyset) = 0$. Then, IVq-ROFCG is show as Equation (18):

$$IVq - ROFCG(a(x_1), a(x_2), \ldots, a(x_n)) =$$

$$< \begin{bmatrix} \prod_{i=1}^{n}(t_a^-(x_{\sigma(i)}))^{\mu(B_{\sigma(i)}) - \mu(B_{\sigma(i-1)})}, \\ \prod_{i=1}^{n}(t_a^+(x_{\sigma(i)}))^{\mu(B_{\sigma(i)}) - \mu(B_{\sigma(i-1)})} \end{bmatrix}, \begin{bmatrix} \sqrt[q]{1 - \prod_{i=1}^{n}(1 - f_a^-(x_{\sigma(i)})^q)^{\mu(B_{\sigma(i)}) - \mu(B_{\sigma(i-1)})}}, \\ \sqrt[q]{1 - \prod_{i=1}^{n}(1 - f_a^+(x_{\sigma(i)})^q)^{\mu(B_{\sigma(i)}) - \mu(B_{\sigma(i-1)})}} \end{bmatrix} > (18)$$

The proof of Theorem 3 is similar to Theorem 1, so the proof is omitted.

**Theorem 4.** Let $a(x_i) = <[t_a^-(x_i), t_a^+(x_i)], [f_a^-(x_i), f_a^+(x_i)]> (i = 1,2, \ldots, n)$ be an IVq-ROFS, $\mu$ is a fuzzy measure on nonempty finite set $X$, IVq-ROFCG satisfies the following four properties:

(1) Idempotency: Assuming there is $a(x) = <[t_a^-(x), t_a^+(x)], [f_a^-(x), f_a^+(x)]>$, which fulfills that $a(x) = a(x_i)(i = 1,2, \cdots, n)$, then IVq-ROFCG$(a(x_1), a(x_2), \cdots, a(x_n)) = a(x)$.

(2) Boundedness: Suppose $a_{min} = <[min(t_{a_i}^-), min(t_{a_i}^+)], [max(f_{a_i}^-), max(f_{a_i}^+)]>$ and $a_{max} = <[max(t_{a_i}^-), max(t_{a_i}^+)], [min(f_{a_i}^-), min(f_{a_i}^+)]>$ are two IVq-ROFNs, then $a_{min} \leq$ IVq-ROFCG$(a(x_1), a(x_2), \ldots, a(x_n)) \leq a_{max}$.

(3) Commutativity: If $(a'(x_1), a'(x_2), \ldots, a'(x_n))$ is the permutation of $(a(x_1), a(x_2), \ldots, a(x_n))$, then the equality IVq-ROFCG$(a(x_1), a(x_2), \ldots, a(x_n)) =$ IVq-ROFCG$(a'(x_1), a'(x_2), \ldots, a'(x_n))$ holds.

(4) Monotonicity: If IV-ROFS $\beta(x_i) = <[t_\beta^-(x_i), t_\beta^+(x_i)], [f_\beta^-(x_i), f_\beta^+(x_i)]> (i = 1,2, \ldots, n)$, for any $\beta(x_i)$, satisfies that $t_{a_i}^- \leq t_{\beta_i}^-, t_{a_i}^+ \leq t_{\beta_i}^+, f_{a_i}^- \geq f_{\beta_i}^-, f_{a_i}^+ \geq f_{\beta_i}^+$, then we have IVq-ROFCG$(a(x_1), a(x_2), \ldots, a(x_n)) \leq$ IVq-ROFCG$(\beta(x_1), \beta(x_2), \ldots, \beta(x_n))$.

The proof of Theorem 4 is similar to Theorem 2, so the proof is omitted.

**Definition 11.** Let $\mu$ be a fuzzy measure on $X$ ($\mu(\emptyset) = 0$) and $a(x_i) = <[t_a^-(x_i), t_a^+(x_i)], [f_a^-(x_i), f_a^+(x_i)]> (i = 1,2, \cdots, n)$ be n IVq-ROFNs.

(1) If $\mu(B \cup C) = \mu(B) + \mu(C)$ for all $B, C \subseteq X$, $B \cap C = \emptyset$, and it is independent for any elements in $X$ that means $\mu(B) = \sum_{x_i \in B} \mu(\{x_i\})$. Then, the IVq-ROFCG transform to interval-valued q-rung orthopair fuzzy weighted Choquet geometric (IVq-ROFWCG) operator, which shows as:

$$IVq - ROFWCG(a(x_1), a(x_2), \ldots, a(x_n)) =$$

$$< \begin{bmatrix} \prod_{i=1}^n (t_a^-(x_{\sigma(i)}))^{\mu(\{x_i\})}, \\ \prod_{i=1}^n (t_a^+(x_{\sigma(i)}))^{\mu(\{x_i\})} \end{bmatrix}, \begin{bmatrix} \sqrt[q]{1 - \prod_{i=1}^n (1 - f_a^-(x_{\sigma(i)})^q)^{\mu(\{x_i\})}}, \\ \sqrt[q]{1 - \prod_{i=1}^n (1 - f_a^+(x_{\sigma(i)})^q)^{\mu(\{x_i\})}} \end{bmatrix} > \quad (19)$$

(2) If for all $B \subseteq X$, there is $\mu(B) = \sum_{i=1}^{|B|} \lambda_i$, we have $\lambda_i = \mu(B_{\sigma(i)}) - \mu(B_{\sigma(i-1)})(i = 1,2, ..., n)$, where $\lambda = (\lambda_1, \lambda_2, ..., \lambda_n)^T (\lambda_n \geq 0)$ and $\sum_{i=1}^n \lambda_i = 1 \ (i = 1,2, ..., n)$. Further, the IVq-ROFCG reduce to interval-valued q-rung orthopair fuzzy order Choquet geometric (IVq-ROFOCG) operator as Equation (20).

$$IVq - ROFOCG(a(x_1), a(x_2), ..., a(x_n)) =$$

$$< \begin{bmatrix} \prod_{i=1}^n (t_a^-(x_{\sigma(i)}))^{\lambda_i}, \\ \prod_{i=1}^n (t_a^+(x_{\sigma(i)}))^{\lambda_i} \end{bmatrix}, \begin{bmatrix} \sqrt[q]{1 - \prod_{i=1}^n (1 - f_a^-(x_{\sigma(i)})^q)^{\lambda_i}}, \\ \sqrt[q]{1 - \prod_{i=1}^n (1 - f_a^+(x_{\sigma(i)})^q)^{\lambda_i}} \end{bmatrix} > \quad (20)$$

(3) If for all $B \subseteq X$, there is $\mu(B) = Q(\sum_{x_i \in B} \mu(\{x_i\}))$ where Q is a basic unit-interval monotonic function, satisfies monotonicity in [0,1] and follows properties: $(i) Q(0) = 0; (ii) Q(1) = 1; (iii) Q(x) \geq Q(y)$ for $x > y$. Then we let $w_i = \mu(B_{\sigma(i)}) - \mu(B_{\sigma(i-1)}) = Q(\sum_{j \leq i} \mu(\{x_{\sigma(i)}\})) - Q(\sum_{j < i} \mu(\{x_{\sigma(i)}\}))(i = 1,2, ..., n)$, where $w = (w_1, w_2, ..., w_n)^T (w_n \geq 0)$ and $\sum_{i=1}^n w_i = 1 \ (i = 1,2, ..., n)$. The IVq-ROFCG change to Equation (21) that is called interval-valued q-rung orthopair fuzzy order weighted Choquet geometric (IVq-ROFOWCG) operator.

$$IVq - ROFOWCG(a(x_1), a(x_2), ..., a(x_n)) =$$

$$< \begin{bmatrix} \prod_{i=1}^n (t_a^-(x_{\sigma(i)}))^{w_i}, \\ \prod_{i=1}^n (t_a^+(x_{\sigma(i)}))^{w_i} \end{bmatrix}, \begin{bmatrix} \sqrt[q]{1 - \prod_{i=1}^n (1 - f_a^-(x_{\sigma(i)})^q)^{w_i}}, \\ \sqrt[q]{1 - \prod_{i=1}^n (1 - f_a^+(x_{\sigma(i)})^q)^{w_i}} \end{bmatrix} > \quad (21)$$

Espcially, if for all $i = 1,2, ..., n$, existing that $\mu(\{x_i\}) = \frac{1}{n}$, then the IVq-ROFOWCG operator reduces to the IVq-ROFOCG operator.

## 4. Group decision-making method based on IVq-ROFCA

In this section we develop the group decision-making method based on IVq-ROFCA. For a MAGDM problem, let $X = \{x_1, x_2, ..., x_n\}$ be the set of alternatives, the set of $C = \{C_1, C_2, ..., C_n\}$ represents attributes, and $\mu(C_j)(j = 1,2, ..., n)$ is the fuzzy measure of attributes. In addition, there is a set of experts $e = \{e_1, e_2, ..., e_t\}$. For each expert, a decision matrix can be obtained according to different alternatives and attribute. For the k-th expert $e_k$, the decision matrix is expressed as: $A^{(k)} = (a_{ij}^{(k)})_{m \times n}$, $a_{ij}^{(k)}$ means the k-th expert's decision value for attribute $j$ in alternative $i$. $a_{ij}^{(k)}$ is an IVq-ROFNs that satisfies $(t_{a_{ij}^{(k)}}^+)^q + (f_{a_{ij}^{(k)}}^+)^q \leq 1 (q \geq 1)$, where the value of q is adapted to the actual situation. Because of the differences between physical dimensions of attributes, the decision matrix is required to be standardized. Correspondingly, $\Omega_1$ is assumed here to represent the benefit type and $\Omega_2$ represent the cost type, and the standardization processing shown as Equations (22), and Equation (23) is the complement operation of fuzzy numbers.

$$r_{ij}^{(k)} = \begin{cases} a_{ij}^{(k)}, a_{ij}^{(k)} \in \Omega_1 \\ (a_{ij}^{(k)})^c, a_{ij}^{(k)} \in \Omega_2 \end{cases} (i = 1,2, ..., m, j = 1,2, ..., n) \quad (22)$$

$$(a_{ij}^{(k)})^c = < \left[f_{a_{ij}^{(k)}}^-, f_{a_{ij}^{(k)}}^+\right], [t_{a_{ij}^{(k)}}^-, t_{a_{ij}^{(k)}}^+] > (i = 1,2, ..., m, j = 1,2, ..., n) \quad (23)$$

The steps of the group decision-making method are as follows:
(1) Deriving the fuzzy measure and the subset of experts and attributes;
(2) According to the evaluation matrix given by experts, choose the appropriate value of q. When the amount of data is small, the value of q can be determined by observation. If the amount of data is large, whether q meets $\left(t_{a_{ij}^{(k)}}^+\right)^q + \left(f_{a_{ij}^{(k)}}^+\right)^q \leq 1 (q \geq 1)$ can be tested and selected by traversal method.

(3) Using Equations (22) and (23) convert the decision matrix $A^{(k)}$ into $A'^{(k)}$;

(4) The IVq-ROFCA is used to aggregate $A'^{(k)} = (a'^{(k)}_{ij})_{m \times n}$ into $R = (r_{ij})_{m \times n}$, and the collection matrix is derived by Equation (24), $r_{ij}^{(t)}$ represents the IVq-ROFNs in the decision matrix given by the t-th expert.

$$r_{ij} = IVq - ROFCA\big(a'^{(1)}_{ij}, a'^{(2)}_{ij}, \ldots, a'^{(t)}_{ij}\big)(i = 1,2,\ldots,m, j = 1,2,\ldots,n) \quad (24)$$

(5) In the second process of aggregation is carried out through IVq-ROFCA, Equation (25) is used to calculate each row in $R$, which will generate the IVq-ROFNs $r_i$ of the corresponding $i$-th row as the result.

$$r_i = IVq - ROFCA(r_{i1}, r_{i2}, \ldots, r_{in})(i = 1,2,\ldots,m) \quad (25)$$

(6) The score value and exact value of $r_i$ are obtained according to Equation (9) and (10), which can further determine the ranking result of alternatives according to the Definition 6 as the ranking of $r_i$.

**5. Case Study on the warning management system for hypertension**

*5.1. Decision result by proposed group decision-making method*

In order to improve the management efficiency of doctors, we plan to develop a daily follow-up warning management system for hypertension. It is decided that the warning level is divided by 5 colors, $x_i(i = 1,2,3,4,5)$, $x_1$ indicates that emergency treatment is required and the resident shall seek treatment immediately, showing red; $x_2$ indicates that treatment is required and residents need treatment in time, showing orange; $x_3$ indicates the requirement for timely follow-up to promote management, showing yellow; $x_4$ that blood pressure management is needed in the future, and the color is blue; $x_5$ indicates that the management service is not needed temporarily, the blood pressure is normal, and the color is green. The warning level is related to various factors of hypertension patients, among which the blood pressure measurements, related diseases, related risk factors, and follow-up time intervals are represented as the attributes, marked by $C = \{C_1, C_2, C_3, C_4\}$. From a managerial perspective, $C_1$, $C_2$, and $C_3$ are benefit types, and $C_4$ is cost type. With different importance, fuzzy measured of $\{C_1, C_2, C_3, C_4\}$ show in Table 2:

**Table 2.** Fuzzy measure of attributes

| Fuzzy measure of $\{C_1, C_2, C_3, C_4\}$ | |
| --- | --- |
| $\mu(\{\emptyset\}) = 0$ | $\mu(\{C_2,C_3\}) = 0.45$ |
| $\mu(\{C_1\}) = 0.2$ | $\mu(\{C_2,C_4\}) = 0.5$ |
| $\mu(\{C_2\}) = 0.2$ | $\mu(\{C_3,C_4\}) = 0.57$ |
| $\mu(\{C_3\}) = 0.25$ | $\mu(\{C_1,C_2,C_3\}) = 0.65$ |
| $\mu(\{C_4\}) = 0.35$ | $\mu(\{C_1,C_2,C_4\}) = 0.77$ |
| $\mu(\{C_1,C_2\}) = 0.42$ | $\mu(\{C_1,C_3,C_4\}) = 0.8$ |
| $\mu(\{C_1,C_3\}) = 0.45$ | $\mu(\{C_2,C_3,C_4\}) = 0.8$ |
| $\mu(\{C_1,C_4\}) = 0.5$ | $\mu(\{C_1,C_2,C_3,C_4\}) = 1.0$ |

Without taking medicine, a 60 years old resident had a measured value and related factors as following: the measured value of systolic blood pressure was 153 mmHg, the measured value of diastolic blood pressure was 98 mmHg, there was no relevant disease record, obesity and family genetic history, and the doctor had no follow-up record. The evaluation value given by three experts $e = \{e_1, e_2, e_3\}$ invited is represented by IVq-ROFNs, and the fuzzy measure of them are $\mu\{e_1\} = \mu\{e_2\} = \mu\{e_3\} = 0.4$, $\mu\{e_1, e_2\} = \mu\{e_2, e_3\} = \mu\{e_1, e_3\} = 0.7$, $\mu\{e_1, e_2, e_3\} = 1$. See Table 3-5 for the evaluation values of the three experts after standardization. We will use the proposed group decision-making method to derive decision result.

**Table 3.** Decision matrix $A'^{(1)}$ by $e_1$

| | $C_1$ | $C_2$ | $C_3$ | $C_4$ |
| --- | --- | --- | --- | --- |
| $x_1$ | < [0.7,0.8],[0.2,0.3] > | < [0.5,0.6], [0.6,0.7] > | < [0.8,0.9], [0.1,0.2] > | < [0.1,0.2],[0.8,0.95] > |

| | $C_1$ | $C_2$ | $C_3$ | $C_4$ |
|---|---|---|---|---|
| $x_2$ | < [0.8,0.9],[0.1,0.2] > | < [0.5,0.55], [0.4,0.5] > | < [0.8,0.9], [0.1,0.15] > | < [0.1,0.2],[0.8,0.95] > |
| $x_3$ | < [0.9,0.95],[0.1,0.2] > | < [0.5,0.6], [0.3,0.4] > | < [0.8,0.85],[0.2,0.3] > | < [0.1,0.2],[0.8,0.9] > |
| $x_4$ | < [0.6,0.7],[0.4,0.5] > | < [0.4,0.5], [0.3,0.4] > | < [0.75,0.8],[0.2,0.3] > | < [0.2,0.3],[0.7,0.8] > |
| $x_5$ | < [0.1,0.2],[0.8,0.9] > | < [0.2,0.3], [0.5,0.6] > | < [0.7,0.85],[0.3,0.4] > | < [0.2,0.3],[0.6,0.7] > |

**Table 4.** Decision matrix $A^{(2)}$ by $e_2$

| | $C_1$ | $C_2$ | $C_3$ | $C_4$ |
|---|---|---|---|---|
| $x_1$ | < [0.6,0.7],[0.2,0.3] > | < [0.6,0.65], [0.4,0.5] > | < [0.8,0.9], [0.1,0.2] > | < [0.1,0.2],[0.8,0.95] > |
| $x_2$ | < [0.9,0.95],[0.1,0.15] > | < [0.5,0.55], [0.4,0.5] > | < [0.8,0.85],[0.1,0.15] > | < [0.1,0.2],[0.8,0.95] > |
| $x_3$ | < [0.85,0.9],[0.1,0.15] > | < [0.5,0.6], [0.3,0.4] > | < [0.8,0.85],[0.2,0.3] > | < [0.1,0.2],[0.85,0.9] > |
| $x_4$ | < [0.4,0.5],[0.3,0.4] > | < [0.4,0.5], [0.3,0.4] > | < [0.75,0.8],[0.2,0.3] > | < [0.2,0.3],[0.85,0.9] > |
| $x_5$ | < [0.1,0.2],[0.8,0.95] > | < [0.2,0.3], [0.5,0.6] > | < [0.7,0.85],[0.3,0.4] > | < [0.2,0.3],[0.7,0.85] > |

**Table 5.** Decision matrix $A'^{(3)}$ by $e_3$

| | $C_1$ | $C_2$ | $C_3$ | $C_4$ |
|---|---|---|---|---|
| $x_1$ | < [0.7,0.8],[0.2,0.3] > | < [0.5,0.6], [0.4,0.5] > | < [0.8,0.9], [0.1,0.2] > | < [0.1,0.2],[0.8,0.95] > |
| $x_2$ | < [0.85,0.9],[0.1,0.2] > | < [0.5,0.55], [0.4,0.5] > | < [0.8,0.85],[0.1,0.15] > | < [0.1,0.2],[0.8,0.95] > |
| $x_3$ | < [0.9,0.95],[0.15,0.2] > | < [0.6,0.7], [0.3,0.4] > | < [0.8,0.85],[0.2,0.3] > | < [0.1,0.2],[0.85,0.9] > |
| $x_4$ | < [0.6,0.7],[0.4,0.5] > | < [0.5,0.6], [0.3,0.4] > | < [0.75,0.85],[0.2,0.3] > | < [0.2,0.3],[0.85,0.9] > |
| $x_5$ | < [0.1,0.2],[0.8,0.9] > | < [0.4,0.5], [0.7,0.8] > | < [0.75,0.8], [0.3,0.4] > | < [0.2,0.3],[0.75,0.85] > |

The value of q is set as 3 via observation, which satisfies the requirements. Then, the collective matrix R can be obtained by IVq-ROFCA, shown as Table 6.

**Table 6.** Collective matrix R

| | $C_1$ | $C_2$ | $C_3$ | $C_4$ |
|---|---|---|---|---|
| $r_1$ | <[0.67,0.78],[0.20,0.30]> | <[0.54,0.62],[0.47,0.57]> | <[0.80,0.90],[0.10,0.20]> | <[0.10,0.20],[0.80,0.95]> |
| $r_2$ | <[0.85,0.92],[0.10,0.18]> | <[0.50,0.55],[0.40,0.50]> | <[0.80,0.87],[0.10,0.15]> | <[0.10,0.20],[0.80,0.95]> |
| $r_3$ | <[0.89,0.94],[0.11,0.18]> | <[0.54,0.64],[0.30,0.40]> | <[0.80,0.85],[0.20,0.30]> | <[0.10,0.20],[0.83,0.90] > |
| $r_4$ | <[0.56,0.66],[0.37,0.47]> | <[0.44,0.54],[0.30,0.40]> | <[0.75,0.82],[0.20,0.30]> | <[0.20,0.30],[0.79,0.86]> |
| $r_5$ | <[0.10,0.20],[0.80,0.91]> | <[0.29,0.39],[0.55,0.65]> | <[0.72,0.84],[0.30,0.40]> | <[0.20,0.30],[0.67,0.79]> |

Calculate each row in R to obtain the comprehensive attribute value of each alternative through IVq-ROFCA, the results show as follow.

$r_1 =< [0.63,0.74], [0.32,0.46] >$
$r_2 =< [0.69,0.77], [0.29,0.39] >$
$r_3 =< [0.72,0.78], [0.32,0.42] >$
$r_4 =< [0.56,0.64], [0.40,0.50] >$
$r_5 =< [0.49,0.60], [0.55,0.66] >$

The score of each alternative can be derived by Equation (9):
$S(r_1) = 0.53, S(r_2) = 0.70, S(r_3) = 0.74,\ S(r_4) = 0.25,\ S(r_5) = -0.12$
Sort the alternatives according to the ranking of scores:

$$S(r_3) > S(r_2) > S(r_1) > S(r_4) > S(r_5)$$

Therefore, the ranking result of alternatives $x_3 > x_2 > x_1 > x_4 > x_5$ can be obtained, and $x_3$ is the optimal selection, which means the doctor is required for timely follow-up to promote management of blood pressure of the patient and receive yellow warning.

*5.2. Paramater analysis*

In order to further verify the influence of the value of q, we conduct the analysis for different values of q in this section, which are separately as 2,3,4 and 5 of this case, and the result obtained is as follows Table 7 and Figure 1 shows.

**Table 7.** Decision result with q=2,3,4,5

| q | Results of IVq-ROFCA | Score of alternatives | Ranking of alternatives |
|---|---|---|---|
| 2 | $r_1 =< [0.60,0.71], [0.32,0.46] >$<br>$r_2 =< [0.65,0.74], [0.29,0.39] >$<br>$r_3 =< [0.69,0.76], [0.32,0.42] >$<br>$r_4 =< [0.53,0.62], [0.40,0.50] >$<br>$r_5 =< [0.44,0.56], [0.55,0.66] >$ | $S(r_1) = 0.55$<br>$S(r_2) = 0.73$<br>$S(r_3) = 0.77$<br>$S(r_4) = 0.26$<br>$S(r_5) = -0.23$ | $x_3 > x_2 > x_1 > x_4 > x_5$ |
| 3 | $r_1 =< [0.63,0.74], [0.32,0.46] >$<br>$r_2 =< [0.69,0.77], [0.29,0.39] >$<br>$r_3 =< [0.72,0.78], [0.32,0.42] >$<br>$r_4 =< [0.56,0.64], [0.40,0.50] >$<br>$r_5 =< [0.49,0.60], [0.55,0.66] >$ | $S(r_1) = 0.53$<br>$S(r_2) = 0.70$<br>$S(r_3) = 0.74$<br>$S(r_4) = 0.52$<br>$S(r_5) = -0.12$ | $x_3 > x_2 > x_1 > x_4 > x_5$ |
| 4 | $r_1 =< [0.65,0.75], [0.32,0.46] >$<br>$r_2 =< [0.71,0.79], [0.29,0.39] >$<br>$r_3 =< [0.74,0.80], [0.32,0.42] >$<br>$r_4 =< [0.58,0.66], [0.40,0.50] >$<br>$r_5 =< [0.53,0.64], [0.55,0.66] >$ | $S(r_1) = 0.44$<br>$S(r_2) = 0.61$<br>$S(r_3) = 0.67$<br>$S(r_4) = 0.21$<br>$S(r_5) = -0.03$ | $x_3 > x_2 > x_1 > x_4 > x_5$ |
| 5 | $r_1 =< [0.67,0.77], [0.32,0.46] >$<br>$r_2 =< [0.72,0.80], [0.29,0.39] >$<br>$r_3 =< [0.75,0.81], [0.32,0.42] >$<br>$r_4 =< [0.60,0.68], [0.40,0.50] >$<br>$r_5 =< [0.56,0.66], [0.55,0.66] >$ | $S(r_1) = 0.38$<br>$S(r_2) = 0.51$<br>$S(r_3) = 0.57$<br>$S(r_4) = 0.18$<br>$S(r_5) = 0.005$ | $x_3 > x_2 > x_1 > x_4 > x_5$ |

It can be seen from Table 7 that the score rankings are always: $S(r_3) > S(r_2) > S(r_1) > S(r_4) > S(r_5)$. That is, the early warning results are all yellow warnings, which are consistent with the results obtained in Table 7, so the decision result does not change due to the variation in the value of q. However, with the increase of q value, it can be easily discovered that scores of the alternatives are declining completely, but show different trends of downward. Considering the practical application, there is not much practical significance of IVq-ROFNs with the large value of q is too large, which often does not exceed 5. Therefore, the value of q will not have an impact on the final decision result.

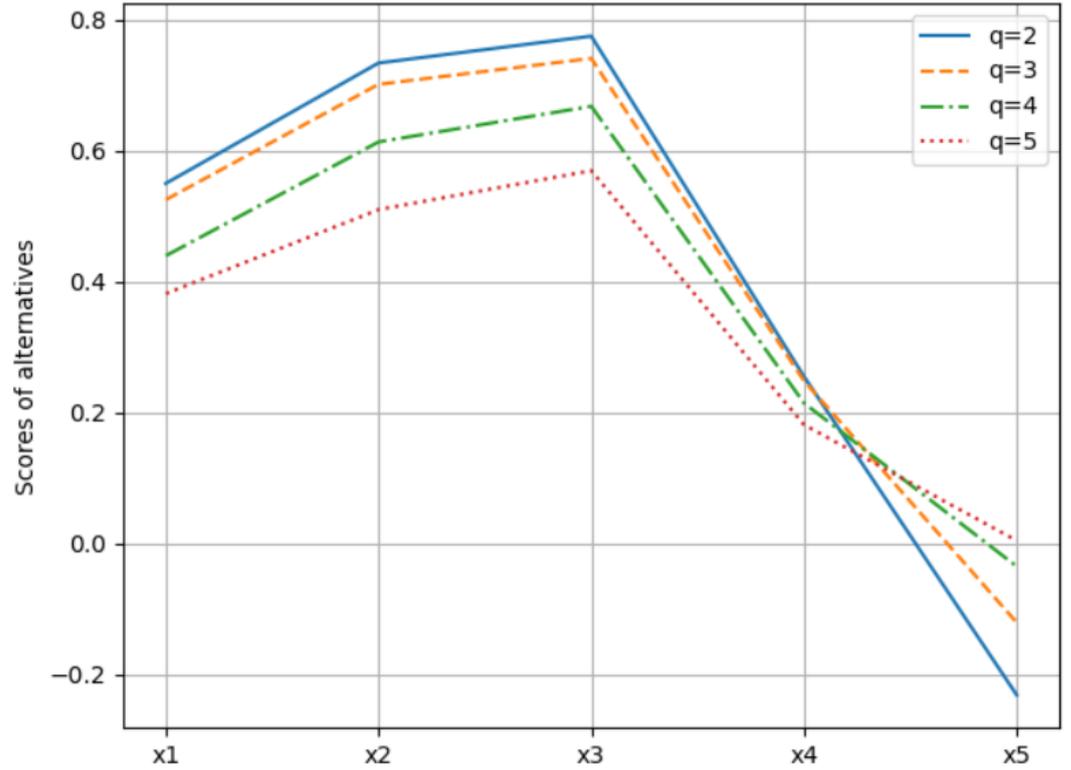

**Figure 1.** Decision result with q=2,3,4,5

*5.3. Comparison analysis*

In this section, the method proposed in this paper is used to solve the case in [26], which is compared with the method given in [26] and [40] and purposed to verify our operators. There are 5 alternatives for 3 experts $e_1$, $e_2$ and $e_3$ to select with 4 attributes $C_1$, $C_2$, $C_3$ and $C_4$, and the evaluation matrices $A^{(1)}$, $A^{(2)}$ and $A^{(3)}$ of interval-valued intuitionistic fuzzy numbers (IVIFNs) given are as Table 8, Table 9 and Table 10. Besides, the fuzzy measures of experts are $\mu\{e_1\} = \mu\{e_2\} = \mu\{e_3\} = 0.4$, $\mu\{e_1, e_2\} = \mu\{e_2, e_3\} = \mu\{e_1, e_3\} = 0.73$, $\mu\{e_1, e_2, e_3\} = 1$, and fuzzy measures of attributes are show in Table 11.

**Table 8.** Decision matrix $A^{(1)}$

|       | $C_1$ | $C_2$ | $C_3$ | $C_4$ |
|-------|-------|-------|-------|-------|
| $x_1$ | < [0.4,0.5],[0.3,0.4] > | < [0.4,0.6],[0.2,0.4] > | < [0.1,0.3],[0.5,0.6] > | < [0.3,0.4],[0.3,0.5] > |
| $x_2$ | < [0.6,0.7],[0.2,0.3] > | < [0.6,0.7],[0.2,0.3] > | < [0.4,0.7],[0.1,0.2] > | < [0.5,0.6],[0.1,0.3] > |
| $x_3$ | < [0.6,0.7],[0.1,0.2] > | < [0.5,0.6],[0.3,0.4] > | < [0.5,0.6],[0.1,0.3] > | < [0.4,0.5],[0.2,0.4] > |
| $x_4$ | < [0.3,0.4],[0.2,0.3] > | < [0.6,0.7],[0.1,0.3] > | < [0.3,0.4],[0.1,0.2] > | < [0.3,0.7],[0.1,0.2] > |
| $x_5$ | < [0.7,0.8],[0.1,0.2] > | < [0.3,0.5],[0.1,0.3] > | < [0.5,0.6],[0.2,0.3] > | < [0.3,0.4],[0.5,0.6] > |

**Table 9.** Decision matrix $A^{(2)}$

|       | $C_1$ | $C_2$ | $C_3$ | $C_4$ |
|-------|-------|-------|-------|-------|
| $x_1$ | < [0.3,0.4],[0.4,0.5] > | < [0.5,0.6],[0.1,0.3] > | < [0.4,0.5],[0.3,0.4] > | < [0.4,0.6],[0.2,0.4] > |
| $x_2$ | < [0.3,0.6],[0.3,0.4] > | < [0.4,0.7],[0.1,0.2] > | < [0.5,0.6],[0.2,0.3] > | < [0.6,0.7],[0.2,0.3] > |
| $x_3$ | < [0.6,0.8],[0.1,0.2] > | < [0.5,0.6],[0.1,0.2] > | < [0.5,0.7],[0.2,0.3] > | < [0.1,0.3],[0.5,0.6] > |
| $x_4$ | < [0.4,0.5],[0.3,0.5] > | < [0.5,0.8],[0.1,0.2] > | < [0.2,0.5],[0.3,0.4] > | < [0.4,0.7],[0.1,0.2] > |
| $x_5$ | < [0.6,0.7],[0.2,0.3] > | < [0.6,0.7],[0.1,0.2] > | < [0.5,0.7],[0.2,0.3] > | < [0.6,0.7],[0.1,0.3] > |

**Table 10.** Decision matrix $A^{(3)}$

|       | $C_1$ | $C_2$ | $C_3$ | $C_4$ |
|-------|-------|-------|-------|-------|
| $x_1$ | < [0.2,0.5],[0.3,0.4] > | < [0.4,0.5],[0.1,0.2] > | < [0.3,0.6],[0.2,0.3] > | < [0.3,0.7],[0.1,0.3] > |
| $x_2$ | < [0.2,0.7],[0.2,0.3] > | < [0.3,0.6],[0.2,0.4] > | < [0.4,0.7],[0.1,0.2] > | < [0.5,0.8],[0.1,0.2] > |
| $x_3$ | < [0.5,0.6],[0.3,0.4] > | < [0.7,0.8],[0.1,0.2] > | < [0.5,0.6],[0.2,0.3] > | < [0.4,0.5],[0.3,0.4] > |
| $x_4$ | < [0.3,0.6],[0.2,0.4] > | < [0.4,0.6],[0.2,0.3] > | < [0.1,0.4],[0.3,0.6] > | < [0.3,0.7],[0.1,0.2] > |
| $x_5$ | < [0.6,0.7],[0.1,0.3] > | < [0.5,0.6],[0.3,0.4] > | < [0.5,0.6],[0.2,0.3] > | < [0.5,0.6],[0.2,0.4] > |

**Table 11.** Fuzzy measures of experts and attributes

| **Fuzzy measures of attributes** ||
|---|---|
| $\mu\{C_1\} = 0.4$ | $\mu\{C_2, C_4\} = 0.43$ |
| $\mu\{C_2\} = 0.25$ | $\mu\{C_3, C_4\} = 0.54$ |
| $\mu\{C_3\} = 0.37$ | $\mu\{C_1, C_2, C_3\} = 0.88$ |
| $\mu\{C_4\} = 0.2$ | $\mu\{C_1, C_2, C_4\} = 0.75$ |
| $\mu\{C_1, C_2\} = 0.6$ | $\mu\{C_1, C_3, C_4\} = 0.84$ |
| $\mu\{C_1, C_3\} = 0.7$ | $\mu\{C_2, C_3, C_4\} = 0.73$ |
| $\mu\{C_1, C_4\} = 0.56$ | $\mu\{C_1, C_2, C_3, C_4\} = 1$ |
| $\mu\{C_2, C_3\} = 0.68$ |  |

For an IVIFS $\hat{a}(x_i) = <[t_{\hat{a}}^-(x_i), t_{\hat{a}}^+(x_i)], [f_{\hat{a}}^-(x_i), f_{\hat{a}}^+(x_i)]> (i = 1,2,\cdots,n)$, the Generalized interval-valued intuitionistic fuzzy geometric aggregation operator (GIIFGA) and interval-valued intuitionistic fuzzy Einstein geometric Choquet integral operator (IVIFEGC) are as follow.

$GIIFGA(\hat{a}(x_1), \hat{a}(x_2), \cdots, \hat{a}(x_n)) =$

$$< \begin{bmatrix} \prod_{i=1}^n (f_{\hat{a}}^-(x_{\sigma(i)}))^{\mu(B_{\sigma(i)})-\mu(B_{\sigma(i+1)})}, \\ \prod_{i=1}^n (f_{\hat{a}}^+(x_{\sigma(i)}))^{\mu(B_{\sigma(i)})-\mu(B_{\sigma(i+1)})} \end{bmatrix}, \begin{bmatrix} 1 - \prod_{i=1}^n \left(1 - t_{\hat{a}}^-(x_{\sigma(i)})\right)^{\mu(B_{\sigma(i)})-\mu(B_{\sigma(i+1)})}, \\ 1 - \prod_{i=1}^n (1 - t_{\hat{a}}^+(x_{\sigma(i)}))^{\mu(B_{\sigma(i)})-\mu(B_{\sigma(i+1)})} \end{bmatrix} > \quad (26)$$

$IVIFEGC(\hat{a}(x_1), \hat{a}(x_2), \cdots, \hat{a}(x_n)) =$

$$< \left[ \frac{2\prod_{j=1}^n t_{\hat{a}}^-(x_{\sigma(i)})_{\sigma(j)}^{\mu(B_{\sigma(i)})-\mu(B_{\sigma(i+1)})}}{\prod_{j=1}^n \left(2 - t_{\hat{a}}^-(x_{\sigma(i)})_{\sigma(j)}\right)^{\mu(B_{\sigma(i)})-\mu(B_{\sigma(i+1)})} + \prod_{j=1}^n t_{\hat{a}}^-(x_{\sigma(i)})_{\sigma(j)}^{\mu(B_{\sigma(i)})-\mu(B_{\sigma(i+1)})}}, \frac{2\prod_{j=1}^n t_{\hat{a}}^+(x_i)_{\sigma(j)}^{\mu(B_{\sigma(i)})-\mu(B_{\sigma(i+1)})}}{\prod_{j=1}^n (2 - t_{\hat{a}}^+(x_i)_{\sigma(j)})^{\mu(B_{\sigma(i)})-\mu(B_{\sigma(i+1)})} + \prod_{j=1}^n t_{\hat{a}}^+(x_i)_{\sigma(j)}^{\mu(B_{\sigma(i)})-\mu(B_{\sigma(i+1)})}} \right],$$

$$\left[ \frac{\prod_{j=1}^n (1+f_{\hat{a}}^-(x_i))^{\mu(B_{\sigma(i)})-\mu(B_{\sigma(i+1)})} - \prod_{j=1}^n (1-f_{\hat{a}}^-(x_i))^{\mu(B_{\sigma(i)})-\mu(B_{\sigma(i+1)})}}{\prod_{j=1}^n (1+f_{\hat{a}}^-(x_i))^{\mu(B_{\sigma(i)})-\mu(B_{\sigma(i+1)})} + \prod_{j=1}^n (1-f_{\hat{a}}^-(x_i))^{\mu(B_{\sigma(i)})-\mu(B_{\sigma(i+1)})}}, \frac{\prod_{j=1}^n (1+f_{\hat{a}}^+(x_i))^{\mu(B_{\sigma(i)})-\mu(B_{\sigma(i+1)})} - \prod_{j=1}^n (1-f_{\hat{a}}^+(x_i))^{\mu(B_{\sigma(i)})-\mu(B_{\sigma(i+1)})}}{\prod_{j=1}^n (1+f_{\hat{a}}^+(x_i))^{\mu(B_{\sigma(i)})-\mu(B_{\sigma(i+1)})} + \prod_{j=1}^n (1-f_{\hat{a}}^+(x_i))^{\mu(B_{\sigma(i)})-\mu(B_{\sigma(i+1)})}} \right] > \quad (27)$$

Where, $(\sigma(1), \sigma(2), \cdots, \sigma(n))$ is a permutation of $(1,2,\ldots n)$ that satisfies $\hat{a}(x_{\sigma(1)}) \leq \hat{a}(x_{\sigma(2)}) \leq \cdots \leq \hat{a}(x_{\sigma(n)})$, $B_{\sigma(i)} = \{x_{\sigma(i)}, x_{\sigma(i+1)}, \ldots, x_{\sigma(n)}\}$, $i = 1,2,\ldots,n$, $B_{\sigma(n+1)} = \emptyset$.

In this case, we set the value of q as 1 and the IVq-ROFNs in the above example are transformed into the IVIFNs, which can be applied to the comparison with exist method. The result of different method obtained show as Table 12 and Figure 2.

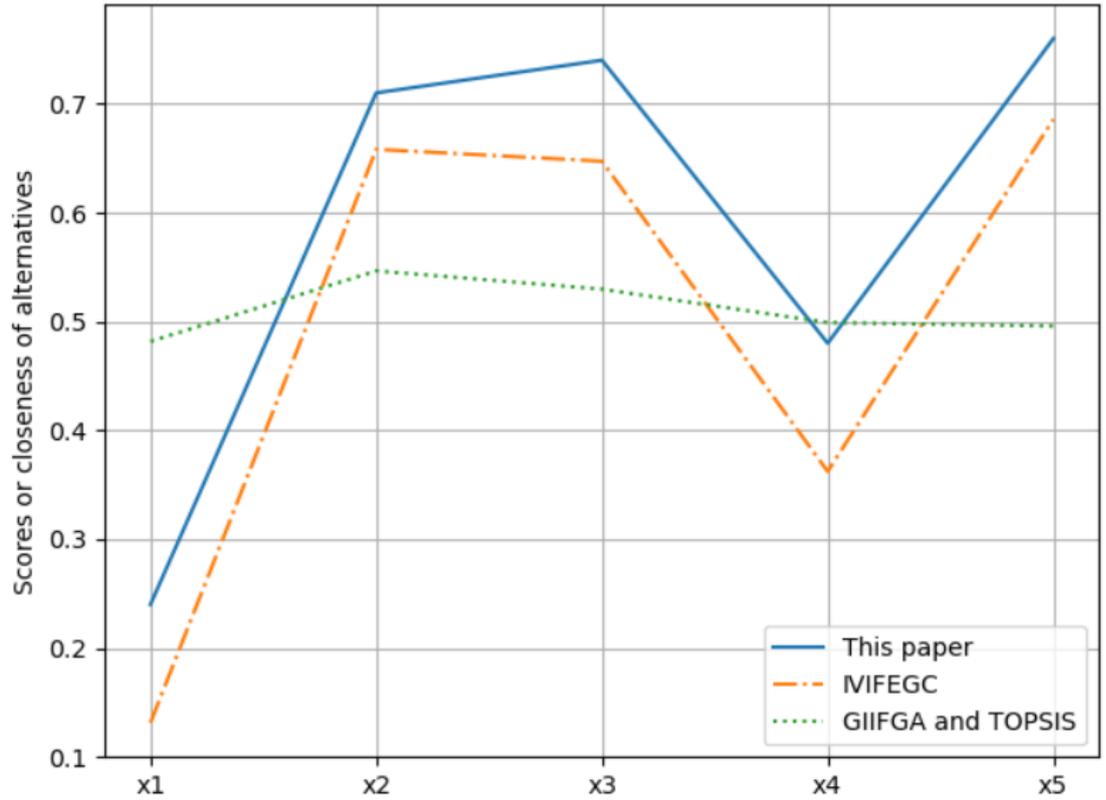

**Figure 2.** Ranking of different methods

**Table 12.** Results by different methods

| Operators | Score or closeness | Alternatives ranking |
|---|---|---|
| **Method proposed in this paper** | Score: <br> $S(x_1) = 0.24$ <br> $S(x_2) = 0.71$ <br> $S(x_3) = 0.74$ <br> $S(x_4) = 0.48$ <br> $S(x_5) = 0.76$ | $x_5 > x_3 > x_2 > x_4 > x_1$ |
| **IVIFEGC[40]** | Score: <br> $S(x_1) = 0.13$ <br> $S(x_2) = 0.66$ <br> $S(x_3) = 0.65$ <br> $S(x_4) = 0.36$ <br> $S(x_5) = 0.69$ | $x_5 > x_2 > x_3 > x_4 > x_1$ |
| **Choquet integral-based TOPSIS (CITOPSIS)[26]** | Closeness: <br> $r(x_1) = 0.4817$ <br> $r(x_2) = 0.5465$ <br> $r(x_3) = 0.5297$ <br> $r(x_4) = 0.4990$ <br> $r(x_5) = 0.4958$ | $x_2 > x_3 > x_4 > x_5 > x_1$ |

It can be seen from Table 12 that in this example, compared with the method in [40], the best and worst choices obtained by the method proposed in this paper are the same, and there are obvious differences in the scores of alternatives. However, there are some differences between $x_2$ and $x_3$, but deviation between them is not large. Using the method proposed in [26] to calculate this example, the decision result is completely different from the other two methods, and the closeness of each alternatives has little difference. The reason is as follows.

In this example, the membership degrees of IVIFNs are generally greater than that of non-membership degrees. In the method investigated by [26], multiple decision matrices are aggregated by CITOPSIS, which will further reduce the difference of the influence of the membership and non-membership of each alternative on the results. Thus, the deviations of the alternatives are not obvious. Contrarily, the method proposed in this paper uses IVq-ROFCA to process the data, which further highlights the degree of expert support for the alternatives. The collective matrices of

GIIFGA and IVq-ROFCA of Table 8-10 are shown as Table 13-14. According to Equation (9), the scores of these two matrices can be derived as Figure 3. From Figure 3, in the matrices show as Table13-14, there is only a small gap between membership and non-membership, except the evaluation of $x_2$ on $C_1$. Correspondingly, it can be seen from Table 8-10 that the experts' evaluation of $x_2$ on $C_1$ gives that existed the span of interval in membership with a wide range, such as [0.2,0.7]. Considering the above, there are differences in the selection of $x_2$ between the method proposed in this paper and [26]. On the other hand, it reflects that the method proposed in this paper will have more advantages when the degree of support of experts for the alternatives is significantly higher than that of opposition

**Table 13.** Collective matrix by GIIFGA

|  | $C_1$ | $C_2$ | $C_3$ | $C_4$ |
|---|---|---|---|---|
| $x_1$ | <[0.3017,0.4645], [0.2685,0.3687]> | <[0.4373,0.5650], [0.1282,0.2983]> | <[0.2452,0.4685], [0.3257,0.4280]> | <[0.3299,0.5720], [0.1911,0.3925]> |
| $x_2$ | <[0.3463,0.5386], [0.2917,0.3925]> | <[0.4353,0.6715], [0.1683,0.2983]> | <[0.4248,0.6715], [0.1282,0.2283]> | <[0.5310,0.7083], [0.1343,0.2616]> |
| $x_3$ | <[0.5712,0.7083], [0.1590,0.2598]> | <[0.5720,0.6732], [0.1590,0.2598]> | <[0.5000,0.6382], [0.1683,0.3000]> | <[0.2751,0.4356], [0.3257,0.4622]> |
| $x_4$ | <[0.3242,0.4996], [0.2283,0.3990]> | <[0.5000,0.7083], [0.1282,0.2616]> | <[0.1951,0.4306], [0.2260,0.3966]> | <[0.3366,0.7000], [0.1000,0.2000]> |
| $x_5$ | <[0.6382,0.7384], [0.1282,0.2616]> | <[0.4685,0.6075], [0.1716,0.2982]> | <[0.5000,0.6382], [0.2000,0.3000]> | <[0.4685,0.5720], [0.2614,0.4280]> |

**Table 14.** Collective matrix by IVq-ROFCA

|  | $C_1$ | $C_2$ | $C_3$ | $C_4$ |
|---|---|---|---|---|
| $x_1$ | <[0.3122,0.4748], [0.3242,0.4248]> | <[0.4288,0.5694], [0.132,0.2944]> | <[0.2575,0.4686], [0.3219,0.4278]> | <[0.3285,0.5722], [0.1871,0.3977]> |
| $x_2$ | <[0.4152,0.6758], [0.2231,0.3242]> | <[0.4632,0.6701], [0.1659,0.2957]> | <[0.4288,0.6758], [0.1206,0.2231]> | <[0.5292,0.7056], [0.1206,0.2624]> |
| $x_3$ | <[0.5694,0.7043], [0.1437,0.2514]> | <[0.5776,0.6818], [0.1552,0.2639]> | <[0.5000,0.6299], [0.1516,0.3000]> | <[0.3306,0.4524], [0.2928,0.4563]> |
| $x_4$ | <[0.3285,0.5004], [0.2231,0.3787]> | <[0.5143,0.7043], [0.1257,0.2689]> | <[0.2116,0.4288], [0.1933,0.3456]> | <[0.3285,0.7000], [0.1000,0.2000]> |
| $x_5$ | <[0.6435,0.7449], [0.1206,0.2551]> | <[0.4614,0.5953], [0.1437,0.2957]> | <[0.5000,0.6299], [0.2000,0.3000]> | <[0.4614,0.5647], [0.2393,0.4353]> |

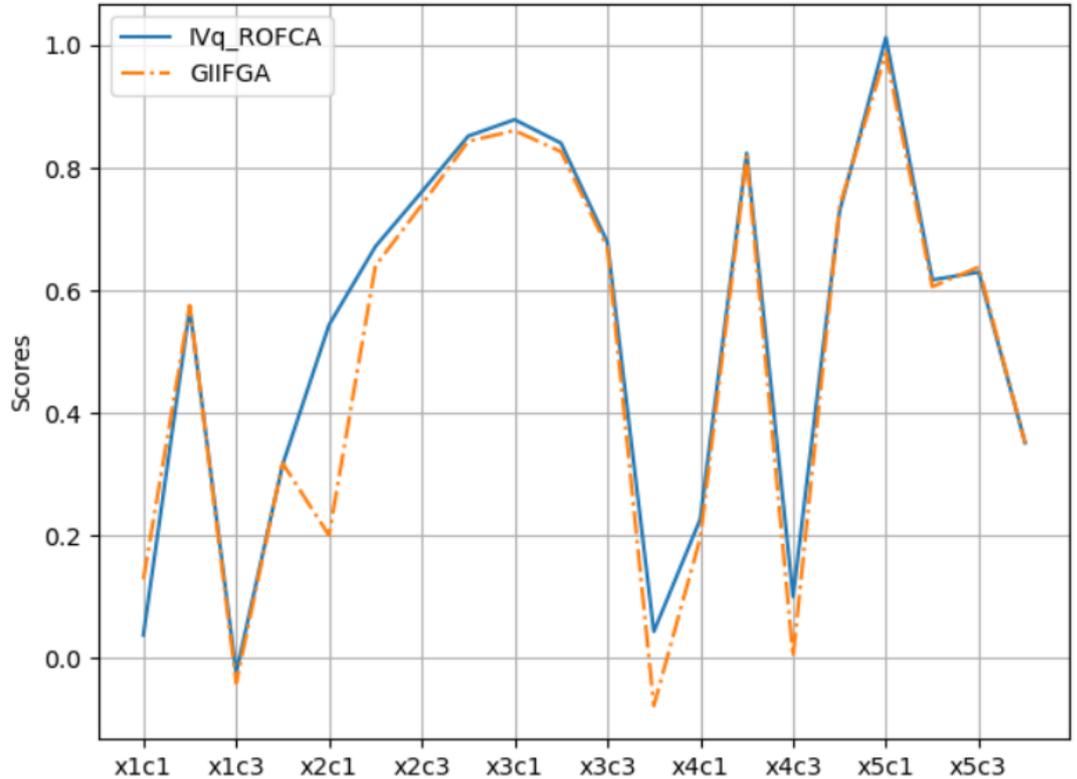
**Figure 3.** Scores of different matrices obtained by GIFFGA and IVq-ROFCA

## 6. Conclusion

This paper proposes several operators of IVq-ROFS based on Choquet integral, and further develops the group decision-making method on this basis. Firstly, four operators of IVq-ROFS based on Choquet integral are investigated, including IVq-ROFCA, IVq-ROFCG, IVq-ROFWCA, IVq-ROFOCA, IVq-ROFOWCA, IVq-ROFWCG, IVq-ROFOCG, IVq-ROFOWCG. Moreover, some properties of them are developed and proofed, such as idempotency, commutativity, monotonicity and boundedness. Then a group decision-making method for MAGDM problems is proposed based on IVq-ROFCA. In this method, the decision matrix of IVq-ROFNs given by experts is collected and then the IVq-ROFNs of different alternatives will be obtained through IVq-ROFCA. Finally, the ranking of alternatives is consistent with the result of corresponding IVq-ROFNs based on score function and exact function. In order to verify the proposed operators and group decision-making method, combined with the case of early-warning model in daily management of hypertension, the results obtained are consistent with the actual management scheme, which is calculated under the evaluation matrix given by medical experts. Meanwhile, it is proved that the operators and group decision-making method proposed in this paper is feasible and effective. Besides, it is proved that the value of q will not affect the decision-making results by set the different value of q under the condition that the q power of membership and non-membership is less than 1. Furthermore, comparing the method proposed in this paper with the existing methods, which proves the rationality of the operators and group decision-making methods proposed in this paper.

Nevertheless, the operators and group decision-making method proposed in this paper still have some limitations. Because the operator proposed in this paper is used to deal with the problems between IVq-ROFNs, which is not able to be applied to other fuzzy environments. Similarly, the group decision-making method proposed in this paper can only solve the MAGDM issues with IVq-ROFNs. Besides, if the decision-making group was relatively large, the degree of consensus between them will affect decision results. Therefore, the method proposed in this paper may be limited in dealing with large-scale group decision-making problems. Further, our future research direction will focus on consensus models of IVq-ROFNs, such as group decision-making method with consistency and consensus of IVq-ROFNs based on Choquet integral. Moreover, considering the broad development prospect of fuzzy theory in the era of big data [41], the integration of group decision-making method, machine learning and big data proposed in this paper is one of our research directions in the future.